# Advancing Head and Neck Cancer Survival Prediction via Multi-Label Learning and Deep Model Interpretation

Meixu Chen[1][0000-0003-3095-1256], Kai Wang[1,2][0000-0003-2155-1445], and Jing Wang[1][0000-0002-8491-4146]

[1] University of Texas Southwestern Medical Center, Dallas, TX
[2] University of Maryland Medical Center, Baltimore, MD
Jing.Wang@UTSouthwestern.edu

**Abstract.** A comprehensive and reliable survival prediction model is of great importance to assist in the personalized management of Head and Neck Cancer (HNC) patient treated with curative Radiation Therapy (RT). In this work, we propose IMLSP, an Interpretable Multi-Label multi-modal deep Survival Prediction framework for predicting multiple HNC survival outcomes simultaneously and provide time-event specific visual explanation of the deep prediction process. We adopt Multi-Task Logistic Regression (MTLR) layers to convert survival prediction from a regression problem to a multi-time point classification task, and to enable predicting of multiple relevant survival outcomes at the same time. We also present Grad-Team, a Gradient-weighted Time-event activation mapping approach specifically developed for deep survival model visual explanation, to generate patient-specific time-to-event activation maps. We evaluate our method with the publicly available RADCURE HNC dataset, where it outperforms the corresponding single-modal models and single-label models on all survival outcomes. The generated activation maps show that the model focuses primarily on the tumor and nodal volumes when making the decision and the volume of interest varies for high- and low-risk patients. We demonstrate that the multi-label learning strategy can improve the learning efficiency and prognostic performance, while the interpretable survival prediction model is promising to help understand the decision-making process of AI and facilitate personalized treatment. The project website can be found at https://github.com/***.

## 1 Introduction

Head and neck cancers (HNC) encompass a spectrum of malignancies affecting the oral cavity, throat, and adjacent regions. For treating HNC patients, radiation therapy is often required for managing the disease [1, 2]. However, the accurate prognosis and personalized treatment of HNC remain challenging in radiation oncology, given the limited number of training data, varied anatomical sites and heterogeneous patient-specific responses to treatment [3-5]. While investigations into patients' electronic health records (EHR) data and radiological image data have shown improved prognostic value compared to the widely used tumor staging based method, most of the prognostic models for RT outcome have predominantly focused on single-task learning frameworks, aiming to predict a specific outcome [6-9]. While these models have provided valuable insights, they often neglect the interconnected nature of different clinical outcomes and the potential for shared representation learning across multiple tasks or labels. In addition, for deep learning-based prediction, scant attention has been given to deep model interpretation analysis to elucidate the decision-making process. These oversights limit the models' ability to harness the full spectrum of available data, potentially obscuring critical insights into the multifaceted nature of cancer progression and treatment response.

Recent advancements in machine learning, particularly in multi-task or multi-label learning, offer a promising method to handle these limitations. By jointly learning to predict multiple related outcomes, these approaches can leverage shared patterns and distinctions among tasks, leading to more robust and generalizable models [10-14]. This is especially pertinent in the context of HNC, where outcomes including overall survival (OS), recurrence free survivals (RFS), and treatment toxicities are crucial in the treatment planning and are routinely documented for post-treatment surveillance to assess treatment response. Notably, for certain HNC RT outcomes, retrospective statistical analyses have revealed strong correlations [15-20].

Outcome prediction models for assisting personalized cancer treatment are high-stake application, where there is a critical need for model transparency and interpretability. Although the integration of deep learning models into prognostic application has demonstrated remarkable success, the complexity of deep neural networks often poses challenges in understanding the decision-making process, limiting their widespread adoption in clinical settings. To address this, recent efforts have explored the application of Gradient-weighted Class Activation Mapping (Grad-CAM) and related techniques in outcome prediction tasks, providing a valuable means to visualize and interpret model decisions [21-23]. These methods offer transparency by highlighting regions of importance in the input data, aiding in understanding where and how the model focuses its attention during classification applications. However, extending these visualization techniques to survival prediction, especially in the context of time-to-event analyses, faces the complexity inherent in framing such problems as regression tasks [24, 25]. This necessitates further development of interpretability tools to effectively unravel the intricacies of survival prediction models.

In this paper, we propose a novel deep framework IMLSP, that integrates multi-label learning approach to simultaneously predict a set of survival outcomes for HNC patients and a deep model interpretation technique to elucidate the decision-making process. Our model is designed to leverage the multimodal data typically available in clinical settings, including patient demographics, tumor characteristics, treatment histories, and planning CT. By doing so, we aim to boost the survival prediction accuracy for multiple outcomes of interest in clinic, provide a more holistic view of patient prognosis, generate patient-specific time-event visual explanation of the deep survival prediction process, and ultimately enabling clinicians to make more informed decisions regarding personalized treatment management. Furthermore, we explore the potential of our model to identify novel examination targets by uncovering patterns and associations across the different tasks and labels, which has the potential to lead to broader understanding of prognosis in HCN radiotherapy.

## 2  Method

### 2.1  Model Overview

Workflow of the proposed IMLSP framework is shown in Fig. 1. A four-layer CNN is used to extract image features, the output is then average pooled and forwarded to one fully connected (FC) layer, concatenated with clinical features. The resulting features are then fed to four Multi-Task Logistic Regression (MTLR) layers for final risk predictions of Overall Survival (OS), Local Failure Free Survival (LFFS), Regional Failure Free Survival (RFFS), and Distant Failure Free Survival (DFFS), respectively. After model training, for a time-event of interest, the corresponding time-event specific activation map is generated via Gradient-weighted Time-event activation mapping (Grad-Team) approach as a visual explanation of the decision-making process of the deep multi-label survival prediction model.

### 2.2  Multi-label Multi-task Logistic Regression Module

MTLR is a method that transforms the survival prediction task from a time-to-event regression problem to a multi-task binary classification problem on discretized time points [7, 26]. This method employs a set of dependent logistic regression units to maintain the consistency of event-free probability prediction. In our IMLSP framework, risk prediction was performed through four separate MTLR paths that received the identical encoded deep imaging features and clinical features from the FC module.

For multi-label survival prediction setting, we divide the time axis into $K$ discrete intervals, with $t_0 = 0$ and $t_K = \infty$, we define $y_s$ as a sequence of binary random variables $(Y_{s,1}, \ldots, Y_{s,K})$, where $Y_{s,k} = \delta(t_{k-1} < t < t_k)$ indicating whether the survival event $s$ occurs in the time interval of $(t_{k-1}, t_k]$, we can write the predicted probability mass function (PMF) as

$$\hat{P}_\Theta(t_{k-1} < t < t_k, S = s|x) = \hat{P}_\Theta\big(y = (Y_{s,1}, \ldots, Y_{s,K-1}), S = s\big|x\big) \quad (1)$$

$$\hat{P}_\Theta(t_{k-1} < t < t_k, S = s|x) = \frac{1}{Z(\Theta_s, x)} \exp\big(\sum_{k=1}^{K-1}(\theta_{s,k}^T x + b_{s,k})y_{s,k}\big) \quad (2)$$

where $\Theta_s = [\theta_{s,k}, b_{s,k}]_{k=1}^{K-1}$ is the parameter set for predicting the risk of outcome $s$, $x$ is the input feature vector, $Z(\Theta_s, x) = \sum_{i=1}^{K} \exp\big(\sum_{k=1}^{K-1}(\theta_{s,k}^T x + b_{s,k})\big)$ is the normalizing constant.

For right censored data for survival label $s$, we can write the PMF by marginalizing over the unseen intervals over the censoring time point $t_{c,s}$:

$$\hat{P}_\Theta(t > t_{c,s}, S = s|x) = \hat{P}_\Theta\big(y = (Y_{s,c}, \ldots, Y_{s,K-1}), S = s\big|x\big) \quad (3)$$

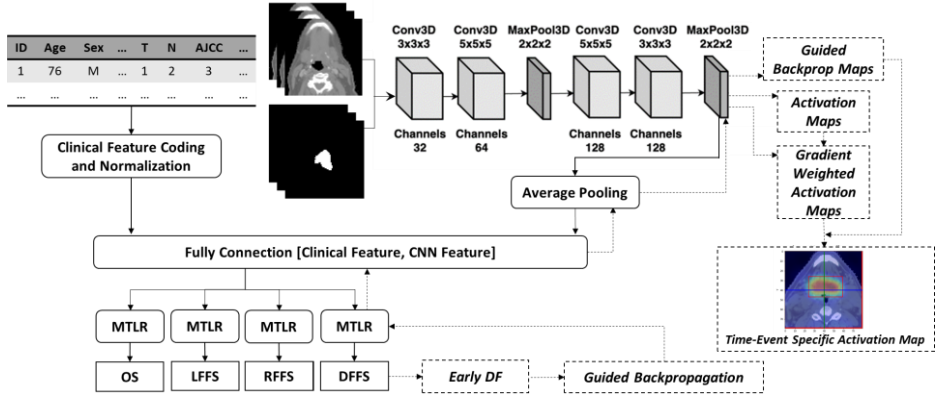

**Fig. 1.** Workflow of the interpretable multi-label multi-modal survival prediction (IMLSP) framework. The left part represents the deep multi-label survival prediction network, and the right part (dished lines) represents the Grad-Team interpretation module for generating patient-specific time-event activation map. Early distant failure (Early DF) is used as an example here.

$$\hat{P}_\Theta(t > t_{c,s}, S = s | x) = \frac{1}{Z(\Theta_s, x)} \sum_{i=c_s}^{K-1} \exp\left(\sum_{k=1}^{K-1}(\theta_{s,k}^T x + b_{s,k}) y_{s,k}\right) \quad (4)$$

For a dataset $Q$ with $R$ patients, and $S$ different survival label recorded, assuming $R_{c,s}$ of whom are right censored for event $s$, and $\lambda_s$ is the weighting factor for event $s$, the multi-label MTLR log-likelihood is given by:

$$L(\Theta, D) = \sum_{s=1}^{S} \lambda_s \sum_{j=1}^{R-R_c} \sum_{k=1}^{K-1} (\theta_{s,k}^T x^j + b_{s,k}) y_{s,k}^j$$

$$+ \sum_{s=1}^{S} \lambda_s \sum_{j=R_c+1}^{R} \sum_{i=1}^{K-1} \delta(t_i > t_{c,s}) \exp\left(\sum_{k=1}^{K-1}(\theta_{s,k}^T x^j + b_{s,k}) y_{s,k}^j\right) \quad (5)$$

$$- \sum_{s=1}^{S} \lambda_s \sum_{j=1}^{R} \log\left(Z(\Theta_s, x^j)\right)$$

We denote $C_\Psi$ as the CNN module with parameter set $\Psi$, $F_\Phi$ as the FC module with parameter set $\Phi$, ∥ as feature concatenation, the predicted survival probability of patient $x$ at time point $t$ for outcome label $s$ obtained using the MTLR module with parameter set $\Theta$ is:

$$\hat{P}_{\Theta,\Phi,\Psi}(t, s, x) = MTLR_s(F_\Phi(x_{clinic} \parallel h_{Vit})) = MTLR_s(F_\Phi(x_{clinic} \parallel C_\Psi(x_{img}))) \quad (6)$$

By replacing the feature vector $x$ with $F_\Phi(x_{clinic} \parallel C_\Psi(x_{img}))$ in formula (5), the loss function $L(\Theta, \Phi, \Psi, D)$ would take all the model parameters. During the model training, we optimize the regularized likelihood $L_{reg}(\Theta, \Phi, \Psi, D) = L(\Theta, \Phi, \Psi, D) + \frac{\beta}{2} \sum_{s=1}^{S} \sum_{k=1}^{K-1} \theta_{s,k}$, to train the whole IMLSP model, where $\beta$ is a regularization hyperparameter [7, 27].

### 2.3   Gradient-Weighted Time-Event Activation Mapping

Similar to the workflow of Grad-CAM, in our Grad-TEAM, we fix the parameters of the proposed model after training, and generate patient-specific time-event activation maps for model interpretation. Given a set of input including imaging data and clinical features, we specify the outcome and time of interest as the interpretation guidance, then we forward propagate the image though the CNN to get the activation map in the last CNN layer. Through time-event-specific computation, we can obtain a guidance vector for encoding the time and event of interest according to the definition in formula (1). The vector is backpropagated to the rectified feature map of interest to compute a gradient weighted activation map for representing where the model looks to make the decision. Finally, we multiply the gradient weighted activation map with guided backpropagated map to get a high-resolution and outcome-specific visualization.

## 3    Experiments and Results

### 3.1    Dataset

We evaluated our method with the RADCURE dataset, which is one of the most extensive HNC imaging datasets publicly available at The Cancer Imaging Archive (TCIA) [28-31]. It contains data from 3346 patients treated with definitive RT at the University Health Network (UHN) in Toronto, Canada between 2005 and 2017. Patient planning CT scans were collected using standard clinical imaging protocols, and the gross tumor volume (GTV) contours were manually generated and reviewed for RT. Furthermore, outcomes data including OS, LFFS, RFFS, and DFFS and clinical characteristics for each patient including demographic, tumor staging, and treatment information were provided. Most of the patients are diagnosed with oropharyngeal cancer, while laryngeal, nasopharyngeal, and hypopharyngeal cancers account for 25%, 12%, and 5% of cases, respectively [31]. Part of RADCURE dataset was employed for a collaborative challenge (RADCURE Challenge) focusing on reproducibility, transparency, and generalizability of modeling for HNC prognosis [29]. To compare our study to the challenge results, we intentionally selected the same cohorts of patients for analysis according to the training and testing patient ID mapping provided by the dataset. After exclusion of patients with missing data, we used 1710 patients for model training and validation and 712 patients for testing. Details of the clinical characteristics of the included patients were summarized in Supplementary Material.

### 3.2    Model Evaluation Metrics

We utilized the C-index to evaluate the accuracy of predicting the life-time risks for different survival outcomes after HNC RT. The 95% confidence interval (CI) of C-index was calculated through 1000-time bootstrapping. Median risk values of different survival outcomes were used as the thresholds to split patients into high- and low-risk groups. Kaplan-Meier survival curves analysis and log-rank test were performed to evaluate the differences in event-free survival distribution. Furthermore, to assess the model's ability to identify high-risk patients for various events at different time points, we employed the area under the receiver operating characteristic curve (AUROC) analysis for prediction of different event at 1-, 2- and 3-years follow-up.

### 3.3 Implementation Details

The CT images and GTV masks are resampled to $1\times 1\times 1$ mm$^3$ and crop into 128×128×64 centered with center of GTV mask. The range of Hounsfield unit (HU) value of CT were truncated to [-500, 500] and normalized to [-1, 1] using min-max normalization. Eleven clinical variables including age, sex, ECOG, smoke status, smoke frequency, T-stage, N-stage, AJCC stage, HPV, chemoradiation, and treatment modality are used as input for our model along with CT patches and GTV mask patches. Clinical variable pre-processing and normalization method are also described in the supplementary.

Our experiments were implemented using PyTorch framework on a 24 GB NVIDIA GeForce RTX 3090 Ti GPU. Uniform weighting was adopted for MTLR losses of different outcomes. Regularization term $\beta$ in MTLR loss function was set as 1 following the default setting. The number of time discretization intervals was selected as 16. Random image 3D rotation and shifting were used for data augmentation. We used AadmW as the optimizer, batch size of 128, and epoch number of 100 for training all the models. The learning rate was set as 0.001 with dynamic learning rate reducing when the multi-label loss stops decreasing on validation cohort (gamma=0.1).

### 3.4 Prediction Performance Comparison

We conducted two sets of comparison experiments to evaluate the prediction accuracy. First, to illustrate the advantages of adopting multiple survival labels simultaneously for outcome prediction, we compared the performance of optimizing single-label survival loss for each outcome individually with the performance when multiple survival losses were optimized together. Single-label survival prediction models for different outcomes were trained following the identical data pre-processing and augmentation procedure as IMLSP model. The best performance reported in the RADCURE challenge, which constructed a deep multitask logistic regression model with EMR features and tumor volume features for OS prediction, were also listed for comparison. Secondly, to identify the contribution of each modality and highlight the benefits of multi-modal feature fusion, we compared the performance of IMLSP model, which fuses clinical data and imaging data for prediction, to models built with clinical features only and imaging data only.

The outcome prediction performance comparison of our proposed model with corresponding single-modal models and single-label models are shown in Tables 1 and 2. The proposed model performs the best on all prediction labels and outperformed the 1st ranked model in the RADCURE Challenge (C-index of OS is 0.800). Kaplan-Meier analysis and ROC analysis are presented in Figs. 2 and 3. In term of C-index, the IMLSP model improved the performance by 0.7% (OS) to 4.9% (RFFS) comparing to single-label models, and 4.3% (OS) to 7.5% (RFFS) comparing to single-modality models. Of note, regional failure (RF) is the outcome of the smallest number of events in RADCURE (99/1710 in training, and 36/712 in testing). When using the median of predicted risk as the threshold to identify high-risk and low-risk patient groups, the log-rank test results showed a significant larger separation between these two (p<0.001 for all comparisons). More details about the results are summarized in the supplementary.

**Table 1.** Life time risk prediction performance comparison of the proposed IMLSP model with clinical-only model, image-only model and single-label models using C-index.

| Model | | C-index | | | |
|---|---|---|---|---|---|
| | | OS | LFFS | RFFS | DFFS |
| Single-Label Multi-Modal | OS Only | 0.817 | 0.648* | 0.553* | 0.711* |
| | LFFS Only | 0.516* | 0.781 | 0.539* | 0.501* |
| | RFFS Only | 0.608* | 0.525* | 0.766 | 0.566* |
| | DFFS Only | 0.490* | 0.546* | 0.377* | 0.762 |
| | RADCURE [29] | 0.801 | NA | NA | NA |
| Multi-Label | Clinical Only | 0.789 | 0.730 | 0.741 | 0.735 |
| | Image Only | 0.777 | 0.764 | 0.748 | 0.725 |
| | Multi-Modal (IMLSP) | **0.823** | **0.799** | **0.804** | **0.781** |

**Table 2.** Event prediction at two-year follow up comparison of the proposed IMLSP model with clinical-only model, image-only model and single-label models using AUROC.

| Model | | AUROC | | | |
|---|---|---|---|---|---|
| | | PD | LF | RF | DF |
| Single-Label Multi-Modal | OS Only | 0.842 | 0.671* | 0.572* | 0.735* |
| | LFFS Only | 0.518* | 0.791 | 0.533* | 0.479* |
| | RFFS Only | 0.622* | 0.543* | 0.787 | 0.572* |
| | DFFS Only | 0.481* | 0.555* | 0.387* | 0.787 |
| | RADCURE [29] | 0.821 | NA | NA | NA |
| Multi-Label | Clinical Only | 0.813 | 0.757 | 0.758 | 0.754 |
| | Image Only | 0.801 | 0.801 | 0.774 | 0.746 |
| | Multi-Modal (IMLSP) | **0.848** | **0.834** | **0.831** | **0.804** |

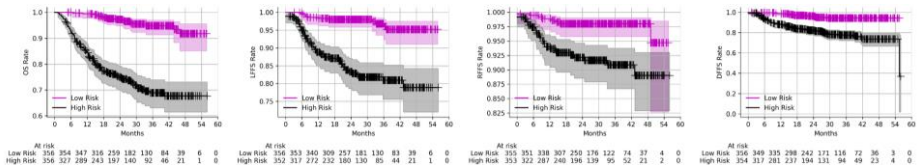

**Fig. 2.** Kaplan-Meier curves of identified low-risk and high-risk patients for OS, LFFS, RFFS, and DFFS using IMLSP (p<0.001 for all). Shaded areas are 95% confidential intervals.

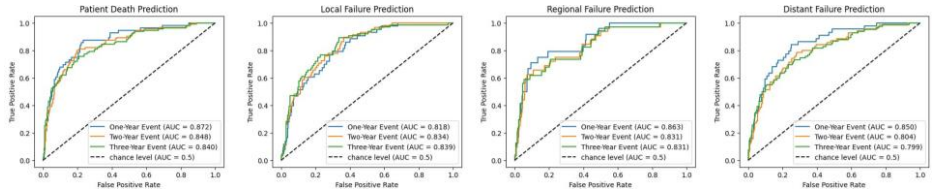

**Fig. 3.** ROC curves for binary prediction of patient death, local, regional and distant failure at 1-, 2-, and 3-years follow-up based on the risk scores from the propose IMLSP model.

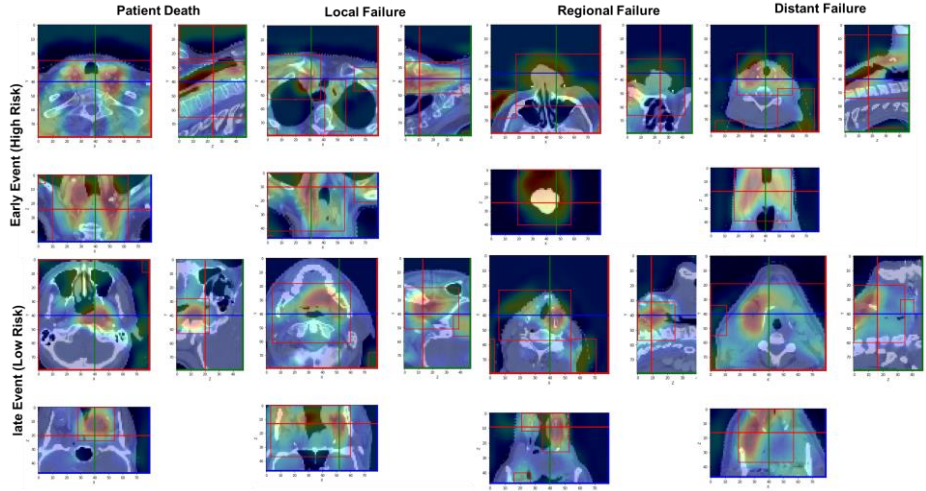

**Fig. 4.** Examples of patient specific time-event activation maps generated from the interpretable deep learning module.

### 3.5　Model Interpretation Ability Analysis

Examples of generated patient-specific time-event activation maps are shown in Fig. 4. The activation maps show that the model focuses primarily on tumor and nodal volumes when making the decision and the volume of interest is larger for high-risk patients (early event patients). For the same patient, the activation maps for different outcome have large overlapping. Comparing to the activation maps generated from single-label survival prediction models (supplementary), the maps generated with IMLSP Grad-Team is more clinically relevant.

## 4　Discussion and Conclusion

In this work, we first propose an interpretable multi-modal multi-label survival prediction model for HNC radiotherapy and a corresponding technique for producing visual explanations of the decision-making process. Our model demonstrates better prognostic ability for all the target outcomes. And the interpretable module is promising to make the decision-making process of AI more transparent and explainable. Our future work will be evaluating the clinical utility of the generated activation maps with physicians, exploring the correlation between imaging data and clinical features, and incorporating more available data modalities, such as genetic data into survival prediction model.

**Acknowledgments.** This study was funded by NIH (R01 CA251792).

**Disclosure of Interests.** The authors have no competing interests to declare that are relevant to the content of this article.

# References

1. Caudell, J.J., et al., *The future of personalised radiotherapy for head and neck cancer.* The Lancet Oncology, 2017. **18**(5): p. e266-e273.
2. Pan, H.Y., et al., *Supply and demand for radiation oncology in the United States: Updated projections for 2015 to 2025.* International Journal of Radiation Oncology* Biology* Physics, 2016. **96**(3): p. 493-500.
3. Beesley, L.J., et al., *Individualized survival prediction for patients with oropharyngeal cancer in the human papillomavirus era.* Cancer, 2019. **125**(1): p. 68-78.
4. Wu, J., et al., *Integrating tumor and nodal imaging characteristics at baseline and mid-treatment computed tomography scans to predict distant metastasis in oropharyngeal cancer treated with concurrent chemoradiotherapy.* International Journal of Radiation Oncology* Biology* Physics, 2019. **104**(4): p. 942-952.
5. Kang, J., et al., *Machine learning approaches for predicting radiation therapy outcomes: a clinician's perspective.* International Journal of Radiation Oncology* Biology* Physics, 2015. **93**(5): p. 1127-1135.
6. Kwan, J.Y.Y., et al., *Radiomic biomarkers to refine risk models for distant metastasis in HPV-related oropharyngeal carcinoma.* International Journal of Radiation Oncology* Biology* Physics, 2018. **102**(4): p. 1107-1116.
7. Kim, S., M. Kazmierski, and B. Haibe-Kains. *Deep-CR MTLR: a Multi-Modal Approach for Cancer Survival Prediction with Competing Risks*. in *Survival Prediction-Algorithms, Challenges and Applications*. 2021. PMLR.
8. Andrearczyk, V., et al., *Overview of the HECKTOR challenge at MICCAI 2022: automatic head and neck tumor segmentation and outcome prediction in PET/CT*, in *Head and Neck Tumor Segmentation and Outcome Prediction: Third Challenge, HECKTOR 2022, Held in Conjunction with MICCAI 2022, Singapore, September 22, 2022, Proceedings*. 2023, Springer. p. 1-30.
9. Chen, M., K. Wang, and J. Wang, *Vision Transformer-Based Multilabel Survival Prediction for Oropharynx Cancer After Radiation Therapy.* International Journal of Radiation Oncology* Biology* Physics, 2023.
10. Jin, C., et al., *Predicting treatment response from longitudinal images using multi-task deep learning.* Nature communications, 2021. **12**(1): p. 1-11.
11. Chen, L., et al., *Attention Guided Lymph Node Malignancy Prediction in Head and Neck Cancer.* International Journal of Radiation Oncology* Biology* Physics, 2021. **110**(4): p. 1171-1179.
12. Zhang, Y. and Q. Yang, *A survey on multi-task learning.* IEEE Transactions on Knowledge and Data Engineering, 2021.
13. Wang, K., et al., *Joint Vestibular Schwannoma Enlargement Prediction and Segmentation Using a Deep Multi‐task Model.* The Laryngoscope, 2022.
14. Zhang, Y., et al. *Multi-label learning from medical plain text with convolutional residual models*. in *Machine Learning for Healthcare Conference*. 2018. PMLR.
15. Ho, A.S., et al., *Decision making in the management of recurrent head and neck cancer.* Head & neck, 2014. **36**(1): p. 144-151.
16. Agrawal, A., et al., *Factors affecting long‐term survival in patients with recurrent head and neck cancer may help define the role of post‐treatment surveillance.* The Laryngoscope, 2009. **119**(11): p. 2135-2140.
17. Eckardt, A., et al., *Recurrent carcinoma of the head and neck: treatment strategies and survival analysis in a 20-year period.* Oral oncology, 2004. **40**(4): p. 427-432.

18. Sun, L., et al., *Association of disease recurrence with survival outcomes in patients with cutaneous squamous cell carcinoma of the head and neck treated with multimodality therapy.* JAMA dermatology, 2019. **155**(4): p. 442-447.
19. Gleich, L.L., et al., *Recurrent advanced (T3 or T4) head and neck squamous cell carcinoma: is salvage possible?* Archives of Otolaryngology–Head & Neck Surgery, 2004. **130**(1): p. 35-38.
20. van der Kamp, M.F., et al., *Predictors for distant metastasis in head and neck cancer, with emphasis on age.* European Archives of Oto-Rhino-Laryngology, 2021. **278**: p. 181-190.
21. Selvaraju, R.R., et al. *Grad-cam: Visual explanations from deep networks via gradient-based localization.* in *Proceedings of the IEEE international conference on computer vision.* 2017.
22. Gotkowski, K., et al. *M3d-CAM: A PyTorch library to generate 3D attention maps for medical deep learning.* in *Bildverarbeitung für die Medizin 2021: Proceedings, German Workshop on Medical Image Computing, Regensburg, March 7-9, 2021.* 2021. Springer.
23. He, T., et al., *Medimlp: using grad-cam to extract crucial variables for lung cancer postoperative complication prediction.* IEEE journal of biomedical and health informatics, 2019. **24**(6): p. 1762-1771.
24. Wang, P., Y. Li, and C.K. Reddy, *Machine learning for survival analysis: A survey.* ACM Computing Surveys (CSUR), 2019. **51**(6): p. 1-36.
25. Lee, C., et al. *Deephit: A deep learning approach to survival analysis with competing risks.* in *Thirty-second AAAI conference on artificial intelligence.* 2018.
26. Yu, C.-N., et al., *Learning patient-specific cancer survival distributions as a sequence of dependent regressors.* Advances in neural information processing systems, 2011. **24**.
27. Fotso, S., *Deep neural networks for survival analysis based on a multi-task framework.* arXiv preprint arXiv:1801.05512, 2018.
28. Clark, K., et al., *The Cancer Imaging Archive (TCIA): maintaining and operating a public information repository.* Journal of digital imaging, 2013. **26**: p. 1045-1057.
29. Kazmierski, M., et al., *Multi-institutional prognostic modelling in head and neck cancer: evaluating impact and generalizability of deep learning and radiomics.* Cancer Research Communications, 2023: p. CRC-22-0152.
30. Welch, M., et al., *Computed tomography images from large head and neck cohort (RADCURE).* 2023, Version.
31. Welch, M.L., et al., *RADCURE: An open‐source head and neck cancer CT dataset for clinical radiation therapy insights.* Medical Physics, 2024.

# Supplementary Material

# Advancing Head and Neck Cancer Survival Prediction via Multi-Label Learning and Deep Model Interpretation

**Supplementary Table 1.** Clinical features included for multi-label survival prediction (IQR: interquartile range; ECOG Performance Status: eastern cooperative oncology group performance status; TN-stage: tumor, Node, Metastasis staging system; AJCC 7th Stage: The seventh edition of the American Joint Committee on Cancer staging system; HPV p16 Status: human papilloma virus p16$^{INK4A}$ expression status; RT: radiation therapy).

| Feature | | Training Cohort Median (IQR)/Number | Testing Cohort Median (IQR)/Number | Description | Normalization/Coding |
|---|---|---|---|---|---|
| Age at Diagnosis | | 61.2 (53.0, 69.8) | 62.7 (54.9-69.3) | numerical, year | z-score normalization |
| Sex | | male: 1361, female: 349 | male: 584, female: 128 | binary/categorical | 1 for male, -1 for female |
| Smoke Status | Cigarettes | 20.0 (0.0-40.0) | 15.0 (0.0-40.0) | numerical | cigarettes day$^{-1}$ × years$_{smoked}$ |
| | Status | current smoker: 164 ex-smoker: 208 non-smoker: 140 unknown: | current smoker: 164 ex-smoker: 208 non-smoker: 140 unknown: | ordinal/categorical | current smoker: -1 ex-smoker: 0 non-smoker: 1 unknown: |
| ECOG Performance Status | | ECOG 0: 1067 ECOG 1: 469 ECOG 2: 139 ECOG >2: 35 | ECOG 0: 408 ECOG 1: 280 ECOG 2: 20 ECOG >2: 4 | ordinal/categorical | ECOG 0: 0 ECOG 1: 1 ECOG 2: 2 ECOG >2: 3 |
| T-stage | | T0: 26 T1: 355 T2: 483 T3: 501 T4: 345 | T0: 10 T1: 151 T2: 218 T3: 198 T4: 135 | ordinal/categorical | T0: 0 T1: 1 T2: 2 T3: 3 T4: 4 |
| N-stage | | N0: 660 N1: 161 N2: 793 N3: 96 | N0: 225 N1: 85 N2: 365 N3: 37 | ordinal/categorical | N0: 0 N1: 1 N2: 2 N3: 3 |
| AJCC 7$^{th}$ Stage | | I: 206 II: 223 III: 337 IVA: 762 IVB: 157 unknown: 25 | I: 67 II: 90 III: 129 IVA: 359 IVB: 58 unknown: 9 | ordinal/categorical | I: 1 II: 2 III: 3 IVA: 4 IVB: 5 Unknown: 0 |
| HPV p16 Status | | positive: 722 unknown: 988 negative: 257 | positive: 455 unknown: 257 negative: 162 | ordinal/categorical | positive: 1 unknown: 0 negative: -1 |
| Chemotherapy | | yes: 687 no: 1023 | yes: 340 no: 372 | binary/categorical | yes: 1 no: -1 |

**Supplementary Table 2.** Summary of follow up and number of events for different survival outcomes in the RADCURE dataset.

| Survival Outcomes | Training Cohort | | Testing Cohort | |
|---|---|---|---|---|
| | Num Event | Median Follow-up (IQR) | Num Event | Median follow up (IQR) |
| Overall Survival | 693 | 4.7 Years (2.7, 5.7) | 114 | 2.4 Years (1.6, 3.2) |
| Local Failure Free Survival | 235 | 4.4 Years (2.4, 5.6) | 67 | 2.3 Years (1.5, 3.2) |
| Regional Failure Free Survival | 99 | 4.7 Years (2.5, 5.7) | 36 | 2.3 Years (1.6, 3.2) |
| Distant Failure Free Survival | 216 | 4.7 Years (2.4, 5.7) | 86 | 2.3 Years (1.5, 3.2) |

**Supplementary Table 3.** Life time risk prediction performance comparison of the proposed multi-modal multi-label model with clinical-only model, image-only model and single-label models. Here C-Index refers to Concordance Index. OS refers to Overall Survival, LFFS refers to Local Failure Free Survival, RFFS refers to Regional Failure Free Survival, and DFFS refers to Distant Failure Free Survival.

| Model | | C-index (95% CI) | | | |
|---|---|---|---|---|---|
| | | OS | LFFS | RFFS | DFFS |
| Multi-Label | Clinical Only | 0.789 (0.732, 0.840) | 0.730 (0.645, 0.807) | 0.741 (0.618, 0.850) | 0.735 (0.661, 0.805) |
| | Image Only | 0.777 (0.723, 0.827) | 0.764 (0.700, 0.825) | 0.748 (0.637, 0.842) | 0.725 (0.657, 0.788) |
| | Multi-Modal | **0.823 (0.771, 0.869)** | **0.799 (0.738, 0.854)** | **0.804 (0.707, 0.887)** | **0.781 (0.718, 0.837)** |
| Single-Label Multi-Modal | OS Only | 0.817 (0.766, 0.865) | NA | NA | NA |
| | LFFS Only | NA | 0.781 (0.721, 0.746) | NA | NA |
| | RFFS Only | NA | NA | 0.766 (0.654, 0.863) | NA |
| | DFFS Only | NA | NA | NA | 0.762 (0.679, 0.835) |
| | RADCURE | 0.801 (0.757–0.842) | NA | NA | NA |

**Supplementary Table 4.** Event prediction at two-year follow up comparison of the proposed multi-modal multi-label model with clinical-only model, image-only model and single-label models. Here AUROC refers to the area under the receiver operating characteristic. PD refers to patient death, LF refers to Local Failure, RF refers to Regional Failure, and DF refers to Distant Failure.

| Model | | AUROC (95% CI) | | | |
|---|---|---|---|---|---|
| | | PD | LF | RF | DF |
| Multi-Label | Clinical Only | 0.813 (0.762, 0.864) | 0.757 (0.687, 0.826) | 0.758 (0.659, 0.857) | 0.754 (0.687, 0.920) |
| | Image Only | 0.801 (0.754, 0.849) | 0.801 (0.748, 0.852) | 0.774 (0.687, 0.860) | 0.746 (0.686, 0.805) |
| | Multi-Modal | 0.848 (0.803, 0.892) | 0.834 (0.786, 0.882) | 0.831 (0.755, 0.907) | 0.804 (0.748, 0.859) |
| Single-Label Multi-Modal | OS Only | 0.842 (0.797, 0.887) | NA | NA | NA |
| | LFFS Only | NA | 0.821 (0.771, 0.870) | NA | NA |
| | RFFS Only | NA | NA | 0.787 (0.700, 0.874) | NA |
| | DFFS Only | NA | NA | NA | 0.787 (0.728, 0.845) |
| | RADCURE | 0.823 (0.777–0.866) | NA | NA | NA |

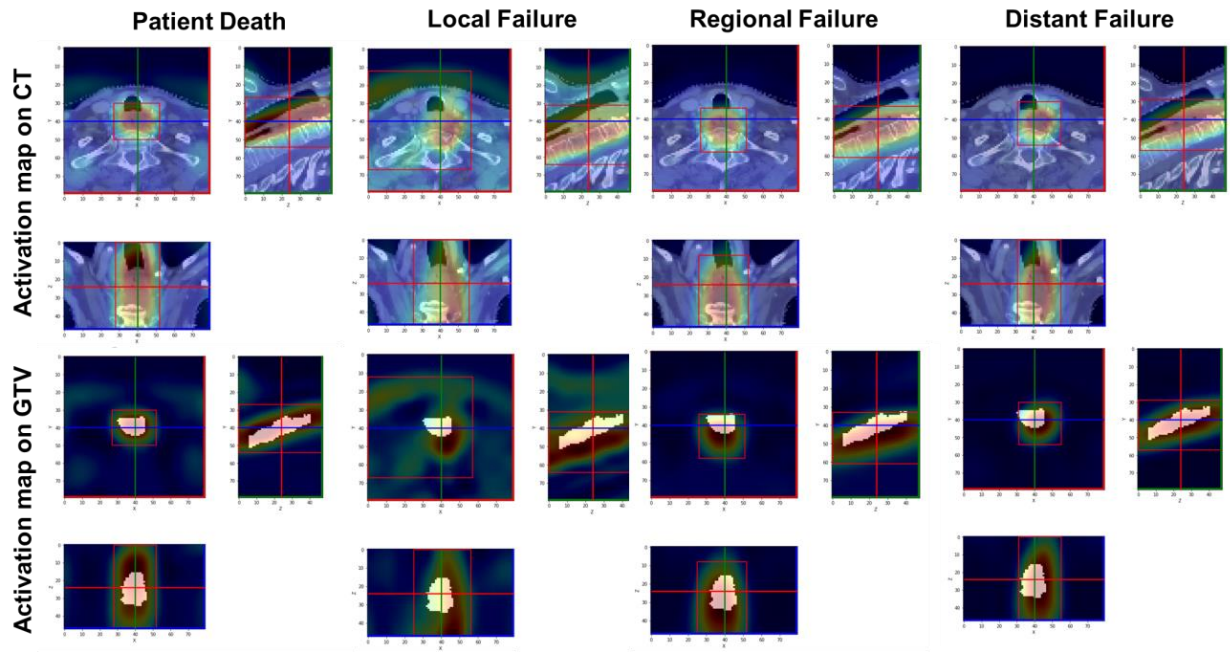

**Supplementary Fig. 1.** Examples of patient specific high risk (early time) activation maps generated from different single label survival prediction models. Activation maps drawn on both the CT image and GTV contours to show where the deep model focus on when making the prediction. These models share the same model structure but are optimized using only the corresponding single survival label MTLR loss.